\documentclass[10pt]{amia}
\usepackage{graphicx}
\usepackage[labelfont=bf]{caption}
\usepackage[superscript,nomove]{cite}
\usepackage{color}
\usepackage{soul}
\usepackage{amsmath}
\usepackage{amssymb}
\usepackage[colorinlistoftodos]{todonotes}
\usepackage[colorlinks=true, allcolors=blue]{hyperref}
\usepackage{subcaption}
\usepackage{float}
\usepackage{algorithm}
\usepackage[noend]{algpseudocode}
\usepackage[autostyle]{csquotes} 
\usepackage[title]{appendix}
\DeclareMathOperator*{\argmax}{arg\,max}

\begin{document}

\title{Improving Sepsis Treatment Strategies by Combining \\Deep and Kernel-Based Reinforcement Learning}

\author{Xuefeng Peng, MSc.$^{1}$, Yi Ding, MSc.$^{2}$, David Wihl, A.L.B.$^{1}$, Omer Gottesman, PhD$^{1}$,\\
Matthieu Komorowski, MD$^{3}$, Li-wei H. Lehman, PhD$^{4}$, Andrew Ross, MSE$^{1}$,\\
Aldo Faisal, PhD$^{3}$, Finale Doshi-Velez, PhD$^{1}$}

\institutes{
    $^1$Harvard University, Paulson School of Engineering and Applied Sciences, Cambridge, MA\\ $^2$Harvard University, T.H. Chan School of Public Health, Cambridge, MA\\$^3$ Imperial College London, London, UK\\ $^4$MIT, Institute for Medical Engineering \& Science, Cambridge, MA\\
}

\maketitle

\noindent{\bf Abstract}

\textit{Sepsis is the leading cause of mortality in the ICU.  It is challenging to manage because individual patients respond differently to treatment.  Thus, tailoring treatment to the individual patient is essential for the best outcomes.  In this paper, we take steps toward this goal by applying a mixture-of-experts framework to personalize sepsis treatment. The mixture model selectively alternates between neighbor-based (kernel) and deep reinforcement learning (DRL) experts depending on patient's current history.  On a large retrospective cohort, this mixture-based approach outperforms physician, kernel only, and DRL-only experts.}

\section*{Introduction}

Sepsis is a medical emergency requiring rapid treatment. \cite{seymour2017time}  It is the cause of 6.0\% of hospital admissions but 15.0\% of hospital mortality.\cite{rhee2017incidence} It is also costly: in 2011 alone, the US spent \$20.3 billion dollars on hospital care for patients with sepsis.\cite{pfuntner2006costs}  Managing sepsis remains challenging, in part because there exists large variation in patient response to existing sepsis management strategies.\cite{waechter2014interaction}  

In this work, we focus on two interventions in the context of sepsis management: intravenous (IV) fluid (adjusted for fluid tonicity) and vasopressors (VP). These two drugs are respectively used to correct the hypovolemia and counteract sepsis-induced vasodilation.  While hypovolemia and vasodilation are common among patients with sepsis, there exists little clinical consensus about when and how these should be treated.\cite{marik2015demise} However, these choices can have large implications for patient mortality:\cite{waechter2014interaction} vasopressors are known to have harmful effects in certain patients, and recent studies have also demonstrated the association between fluid-overload and negative outcomes in the ICUs.\cite{kelm2015}  

Thus, it is essential to identify ways to personalize treatment.  The availability of large observational critical care data sets \cite{johnson2016mimic} has made it possible to hypothesize improved sepsis management strategies, and prior studies \cite{DBLP:journals/corr/RaghuKCSG17,komorowski2018intensive} have used this resource to suggest optimal treatment strategies for patients with sepsis.  
As with those earlier works, we personalize strategies by using reinforcement learning (RL), a technique for optimizing sequences of decisions given patient context.  However, we use a mixture-of-experts approach to combine two very different RL techniques with very different strengths---a model-free deep RL approach (DRL) and a model-based kernel RL approach (KRL)---to improve the quality of the recommended treatment policy.  Specifically, our work extends prior efforts in three important ways:
\begin{enumerate}
\item \textit{Recurrent encoding of the patient's history.}  To date, work in this domain has assumed that the patient's current measurements are sufficient to summarize their history.  To retain potentially decision-relevant information from the patient's past, we use a recurrent autoencoder to represent the patient's entire history.

\item \textit{Safe-guards on the Deep RL.} DRL-based approaches can be particularly poor at extrapolating, and even in areas of dense data, they can suggest non-sensical actions.  We explicitly restrict the DRL to only suggest actions commonly taken by clinicians, moving us toward more clinically-credible policies.

\item \textit{Combining DRL with Kernel RL.} Finally, we use a mixture-of-experts (MoE) to combine the restricted DRL approach with a kernel RL approach selectively based on the context.  DRL is more flexible but can be prone to various pathologies; KRL is guaranteed to stay close to the data but as a result can extrapolate poorly.  The MoE combines their strengths.

\end{enumerate}

\section*{Related Work}

Reinforcement Learning has been applied to a number of applications in healthcare, ranging from emergency decision support,\cite{thapa2005agent} treating malaria, \cite{rajpurkar2017malaria} and managing HIV.\cite{parbhoo2014reinforcement}  Prasad et al.\cite{prasad2017reinforcement} use RL to identify when to the wean patients from mechanical ventilation in ICUs. 

With respect to fluid and vaospressor use in sepsis management, Komorowski et al.\cite{komorowski2018intensive} model a discrete Markov decision process from data and then utilize it to learn a treatment strategy.  Raghu et al.\cite{DBLP:journals/corr/RaghuKCSG17} extend this work by considering a much more expressive continuous representation of patient state.  They use a traditional, non-recurrent autoencoder to first compress measurements from each time step into a continuous state representation, and then they learn a mapping from the state representation to an appropriate treatment via a Dueling Double-Deep Q Network (Dueling DDQN).  Our work uses an even more expressive state representation that represents the patient's entire history, and we also add safe-guards against inappropriate actions and develop richer policies through our mixture of experts.

The mixture of experts aspect of our work builds from ideas developed by Parbhoo et al.\cite{parbhoo2017combining} in the context of HIV management.  In their case, they switch between a kernel-based policy and a discrete Bayesian Partially Observable Markov Decision Process (POMDP). We follow the idea of combining experts, but use the DDQN as an expert rather than a discrete POMDP, as Raghu et al.\cite{DBLP:journals/corr/RaghuKCSG17} have already demonstrated that a continuous expressive state representation is valuable for the sepsis management task.  Because we use a recurrent encoding to summarize the entire patient history, our state can be thought of as a sufficient statistic, much like the POMDP belief state in Parbhoo et al.\cite{parbhoo2017combining}

\section*{Background}

The reinforcement learning framework models a sequence of decisions as an agent interacting with an environment over time.  At each time step $t$, the RL agent observes a state $s$ from the state space $S$ and selects an action $a$ from the action space $A$ based on some policy $\pi(s,a)$, which assigns a probability to action in each state.  Upon taking the action, the agent receives some reward $r$ and transitions to a new state $s^\prime$.  The agent's goal is to maximize their expected longterm discounted return $\mathbb{E} [\sum_t \gamma^t r_t]$.  The optimal value function is defined as $V^{*}(s)=\max_{\pi} \mathbb{E} [\sum_t \gamma^t r_t]|s_0=s, \pi]$, and the optimal state-action value function $Q^{*}(s,a)=\max_{\pi}\mathbb{E} [\sum_t \gamma^t r_t|s_0 = s,a_0 = a,\pi]$.  The latter satisfies the Bellman equation $Q^{*}(s,a)= r(s,a)+\gamma \max_{a^{'}} \mathbb{E}[Q^{*}(s^{'},a^{'})]$, where $\gamma$ is the discount factor determines the trade-off between immediate and future rewards. Q-learning methods aim to learn an optimal policy by minimizing the temporal difference (TD) error, defined as $r(s,a) + \gamma Q^{*}(s^{'}, a^{'}) - Q^{*}(s,a)$.

\section*{Cohort and Data Processing}

\textit{Cohort.} We used the same patient set as in Raghu et al. \cite{DBLP:journals/corr/RaghuKCSG17} which applied the Sepsis-3 
criteria to the Multi-parameter Intelligent Monitoring in Intensive Care (MIMIC-III v1.4) database. \cite{johnson2016mimic} Our cohort consisted of 15,415 adults (age range of 18 to 91), summarized in Table \ref{table:demographics}. 

\begin{table}[h]
\centering
\caption{Comparison of cohort statistics for subjects that fulfilled the Sepsis-3 criteria}
\label{table:demographics}
\begin{tabular}{|l|r|r|r|}
\hline
 & \% Female & Mean Age & Total Population \\
 \hline
 Survivors & 44.1\% & 63.9 &13,535 \\
 \hline
 Non-survivors & 44.3\% & 67.4 & 1,880 \\
 \hline
\end{tabular}
\end{table}

\textit{Cleaning and Preprocessing.}
As in Raghu et al. \cite{DBLP:journals/corr/RaghuKCSG17}, patient histories were partitioned into 4-hour windows each containing fifty attributes, ranging from vitals (e.g. heart rate, mean blood pressure) to clinician assessments of the patient's conditions (e.g. sequential organ failure assessment (SOFA) score).  Patients with missing values were excluded.  The observations, which range from demographics, lab values and vital signs, all have different scales.   Following Raghu et al., \cite{DBLP:journals/corr/RaghuKCSG17} we performed log transformations of observations with large values and standardized the remaining values. After the standardization and log transformation, all values were rescaled into $[0-1]$.  The details of attributes and preprocessing is shown in Table \ref{table:preprocessing}.The data set was split into a fixed 75\% training and validation set and a 25\% test set. 

\begin{table}[H]
\centering
\caption{Physiological attributes, treatments and corresponding preprocessing methods}
\label{table:preprocessing}
\begin{tabular}{|l|l|}
\hline
Preprocessing   & Attributes \\ \hline
Standardization & \begin{tabular}[c]
{@{}l@{}}\texttt{age,Weight\_kg,GCS,HR,SysBP,MeanBP,DiaBP,RR,Temp\_C,FiO2\_1,}\\ 
  \texttt{Potassium,Sodium,Chloride,Glucose,Magnesium,Calcium,Hb,}\\ 
  \texttt{WBC\_count,Platelets\_count,PTT,PT,Arterial\_pH,paO2,paCO2,}\\ 
  \texttt{Arterial\_BE,HCO3,Arterial\_lactate,SOFA,SIRS,Shock\_Index,PaO2\_FiO2,}\\ 
  \texttt{cumulated\_balance\_tev, Elixhauser, Albumin, CO2\_mEqL, Ionised\_Ca}
  \end{tabular} \\ 
  \hline
  Log transformation    & \begin{tabular}[c]
    {@{}l@{}}\texttt{max\_dose\_vaso,SpO2,BUN,Creatinine,SGOT,SGPT,Total\_bili,}\\
    \texttt{INR,input\_total\_tev,input\_4hourly\_tev,output\_total,output\_4hourly}
    \end{tabular}                             \\ \hline
\end{tabular}
\end{table}

\textit{Treatment Discretization.} In this work, we focus on administrating two drugs: intravenous (IV) fluid and vasopressor (VP). In the cohort, the usage of IV and VP for each patient are recorded at each 4-hour window.  Following Raghu et al.\cite{DBLP:journals/corr/RaghuKCSG17}, the dosages for each drug are discretized into 5 bins, resulting in a $5\times5$ action space indexed from 0 to 24. Note that the first action ($a=0$) means ``no action"---neither IV nor VP are prescribed.

\section*{Method Overview}
Applying RL to the sepsis management problem involves several pieces.  The first is defining our inputs and treatments (sections above).  Next we describe how we compress patient histories into a state via a recurrent autoencoder, how we attribute rewards to each state, and also how we determine the quality of some policy given observational data.  With these pieces in place, we finally describe how to derive treatment policies that optimize for our rewards, including our mixture-of-experts (MoE) approach. 

\subsection*{Compressing Patient Histories}

Prior efforts  \cite{komorowski2018intensive,DBLP:journals/corr/RaghuKCSG17} assumed that the patient's current measures were sufficient to summarize their history; however, past and trend patient information is often valuable to deciding the appropriate course of action. To capture more of this temporal information, we encoded patient states recurrently using an LSTM autoencoder representing the cumulative history for each patient.  The LSTM had a single layer of 128 hidden units for both encoder and decoder---resulting in a state $s$ that consisted of 128 real-valued vector. The autoencoder was trained to minimize the reconstruction loss (MSE) between the original measurements and the decoded measurements. We trained the model with mini-batches of 128 and the Adam optimizer for 50 epochs until its convergence.

\subsection*{Reward Formulation}
\label{subsec:rewardformulation}

Broadly, we are interested in reducing mortality among patients with sepsis.  However, mortality is a challenging objective to optimize for because it is only observed after a long sequence of decisions; it can be hard to ascertain which action was responsible for a good or bad outcome.  Thus, for the purposes of our training, we introduce an \emph{intermediate} reward that can give a preliminary signal for whether our sequence of treatment decisions is likely to reduce mortality.  

Specifically, we first train a regressor that predicts the probability of mortality given a patient's current observations (implications of this choice in \nameref{section:discussion}).  Next, we define the reward as the change in the negative mortality log-odds of mortality between the current observations and the next observations.  (Log-odds were used because the probabilities of mortality actually vary over a relatively small range.)  Let $f(o)$ be the probability of mortality given current observations $o$.  Then we define the reward $r(o,a,o^\prime)$ as 
\begin{align}
\label{eqn:network}
r(o,a,o^\prime) = - \log \frac{f(o^\prime)}{1-f(o^\prime)} f(o^\prime) + \log \frac{f(o)}{1-f(o)}
\end{align}
For a sense of scale, among those over 186K patient state transitions from both training and testing sets, the rewards are in the interval $[-3, 3]$.

\begin{figure}[H]
\centering
\includegraphics[width=0.4\linewidth]{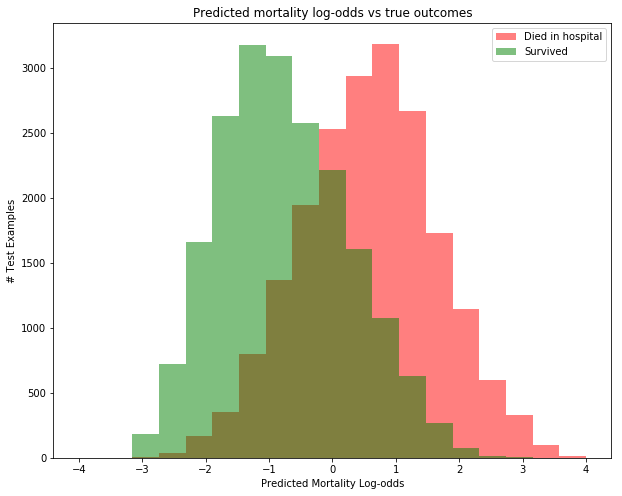}
\caption{\label{fig:mortality-log-odds} Mortality log-odds distribution for mortality and survivor classes.}
\end{figure}

The mortality predictor $f(o)$ itself was a two-layer neural network with $64$ and $32$ units for each layer and L1-regularized gradients (see Ross et al. \cite{nips2017timl} for details). The L1 regularization encourages sparse local linear approximations, and thus makes its behaviors more interpretable.  We resampled balanced batches (between survivors and non-survivors) during training with batch sizes of $128$ observations for $50$ epochs, and our predictor $f(o)$ achieved a test accuracy of $73.1\%$. Figure \ref{fig:mortality-log-odds} shows the log-odds distributions for each class.

\subsection*{Off-Policy Evaluation via the WDR Estimator}
\label{subsec:wdr}
The natural question is how to evaluate the quality of a proposed policy $\pi_e$ given only retrospective data collected according to a clinician policy $\pi_b$. The weighted doubly robust (WDR) estimator\cite{thomas2016data} is widely used for off-policy evaluation in RL. It uses estimated value $\hat{V}$ and action-value $\hat{Q}$ functions as control variates to reduce the variance of the off-policy estimation. In the following, we used the value function of the DRL policy as our control variate; we also explored using the value of the clinician policy, estimated as the mean, instead of the max, of the DRL Q-values over action space.

\begin{align}
\label{eq:wdr}
\text{WDR}(D) := \sum_{i=1}^I\sum_{t=0}^{T} \gamma^tw_i^tr_t^{H_i} -  \sum_{i=1}^I\sum_{t=0}^{T}\gamma^t(w_t^i\hat{Q}^{\pi_e}(S_t^{H_i}, A_t^{H_i}) - w_{t-1}^i\hat{V}^{\pi_e}(S_t^{H_i}))
\end{align}
Here, $I$ is the number of patients, $t$ is the time step, and $H_{i}$ refers to the $i_{th}$ patient's ICU-stay state trajectory. The importance weight of the state of the patient $i$ at time step $t$ is defined as $w^{i}_{t}=\frac{\rho^{i}_{t}}{\sum^{I}_{j=1}\rho^{j}_{t}}$, where $\rho^{i}_{t} = \prod^{t}_{t'=0}\frac{\pi_{e}(A^{H_i}_{t'}|S^{H_i}_{t'})}{\pi_{b}(A^{H_i}_{t'}|S^{H_i}_{t'})}$. The reward $r$, value function $\hat{V}$, and action-value function $\hat{Q}$ are as defined above.  Finally, we estimate the clinician policy $\pi_b$ as the empirical distribution over actions of the 300 neighbors in the training set with states closest to $s$, as research shows that clinicians typically make decisions based on their experience treating similar patients \cite{norman2005research}.

\section*{Deriving Policies}
\label{subsec:dp}
With a means of representing patient history, rewards, and a metric for evaluating policies, we can now start optimizing treatment strategies.  Below we describe each expert---the DRL and the KRL---and how we combine them.

\begin{figure}[H]
\centering
\includegraphics[width=0.7\textwidth]{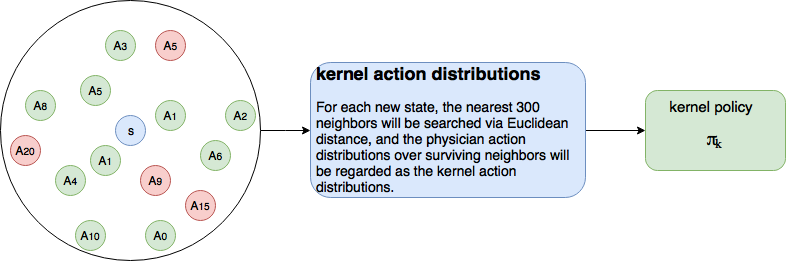}
\caption{\label{fig:kernel_policy} The circle in the left shows an example of the neighborhoods of a new state $s$, red and green marks the mortality and surviving states respectively, and each of these states is associated with a physician action $A_{i}$.}
\end{figure}

\textit{Kernel Policy.} One simple way to derive a treatment decision rule is to look at the nearest neighbors to the current state $s$, identify the survivors, and choose actions that correspond to the distribution of treatments performed on these nearby survivors (see cartoon in Figure \ref{fig:kernel_policy}).  Specifically, we 
\begin{enumerate}
\setlength{\itemsep}{0pt}
\item Get the encoding $s$ from the patient's history $h$ up to time $t$ via the LSTM autoencoder.
\item Search \textit{k} nearest neighbors in the training set in this encoded representation space using Euclidean distance.
\item The kernel policy $\pi_k$ is the distribution of actions taken over the surviving nearest neighbors.
\end{enumerate}
We cross-validated the \textit{k} values ranging from 200 to 500 using WDR, and proceeded with $k=300$. 

\textit{DQN Policy.} Double DQN (DDQN) with dueling structure\cite{wang2015dueling}, which is a variant of DQN, has been applied to derive a policy that outperforms the physician policy\cite{DBLP:journals/corr/RaghuKCSG17}. The structure of dueling DDQN is particularly suitable for sepsis treatment strategy learning, as it differentiates the value function $V$ into the value of the patient's underlying physiological condition, called the \textit{Value} stream, and the value of the treatment given, called the \textit{Advantage} stream.  

We train the dueling DDQN for 200,000 steps with
$batch\,size=30$ to minimize the TD-error. At each given state, the agent is trained to take an action with the highest Q-value, in order to achieve the ultimate goal of improving the overall survival rate.  To stabilize the training process and improve the performance, we applied regularization term $\lambda$ to the Q-network loss to penalize output Q-values which exceeded the maximum observed rewards $r_{max}=3$ 


and used prioritized experience replay\cite{schaul2015prioritized} to balance the training sets with high-value and high-error states.
\begin{equation}
\label{eq:qnetwork}
\textit{\L}(\theta) = E[(Q_{double-target}-Q(s,a;\theta))^{2}] + \lambda\,\max(|Q(s,a;\theta)-r_{max}|, 0)
\end{equation}
where 
\begin{equation}
Q_{double-target} = r + \gamma Q(s,\argmax_{a^\prime} Q(s,a^\prime;\theta^\prime))
\end{equation}
and $\theta, \theta'$ are parameters of the DQN networks.

Finally, given a set of action-values $Q(s,a)$, we must still define a policy $\pi_d$.  Typically, these actions are chosen by the $\max Q$-value, but this ignores the fact that two actions may have very similar values---and given the limitations of our learning, it may not be possible which is actually the best.  We define the DRL policy $\pi_d$ as the softmax of the action-value or \textit{Advantage} stream, giving higher probability to actions with higher values but not forcing us to take (what might be a brittle) best value.

\section*{Mixture-of-Experts (MoE)}
\begin{figure}[H]
\centering
\includegraphics[width=0.7\textwidth]{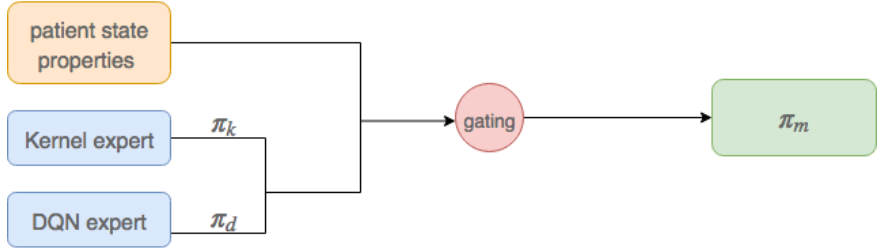}
\caption{\label{fig:moe} The architecture of MoE, it produces a mixed policy via combining kernel.}
\end{figure}

The two approaches above---KRL and DRL---have different strengths.  For patient states which are atypical, i.e. farther Euclidean distance away from any neighbors, the kernel policy may end up relying on neighbors that are not really that similar to the patient.  In contrast, DQN, in trying to fit a value function to the whole state space at once, may still underfit in regions with plentiful data.  Our mixture-of-experts (MoE) uses properties of the patient's current state, and the relationship between the patient's current state and states observed during training, to switch between the kernel and DQN policies (see Figure \ref{fig:moe} for a cartoon).

\textit{Action Restriction.}
The DDQN, as a complex function approximator, comes with relatively few guarantees.  Sometimes, it can place high value on actions that were rarely or never performed by clinicians.  To safe guard against these rare (and likely dangerous) actions, we restricted the DDQN to actions only taken more than 1\% of the time by the physicians among its 300 nearest neighbors. Specifically, let $\pi_{d}(s,a)$ be the DDQN policy, and $\pi_{b}(s,a)$ be the physician policy.  If $\pi_{b}(s,a) < 0.01$, then we set $\pi_{d}(s,a) \leftarrow 0$ and then normalize $\pi_d(s,a)$ to be a valid probability distribution.  (We note that the kernel policy, which is derived directly from clinician actions, cannot deviate in this way; once we restrict the DDQN actions, the MoE will also never take rare actions.) 

\textit{Choice of Gating Function.}
We examined several medical sources \cite{jones2009sequential,beier2011elevation,tamion2010albumin} to determine which features might be most useful for selecting  between experts.  Our final set of features were: age, Elixhauser, SOFA, $FiO_{2}$, BUN, GCS, Albumin, trajectory length, and max distance from neighbors.  For our MoE gating function we combined these features $x$ linearly via weights $w$, along with a bias term $b$, and passed them through a logit to get the probability of choosing each policy: 
\begin{align}
\label{eq:moe}
p_k = \text{sigmoid}(w \cdot x + b) \quad \textrm{and} \quad p_{d} = 1 - p_{k}
\end{align}
where $p_{k}$ and $p_{d}$ denote the assigned probability for choosing the kernel and DQN policy respectively. 

\textit{Optimizing the Gating Function.}
Of course, the core question is how to choose the gating parameters so as to maximize long-term rewards.  Given a set of weights $w$ and bias $b$, the MoE policy $\pi_m$ is defined as $\pi_{m}(s,a)=p_{k}\pi_{k}(s,a) + p_{d}\pi_{d}(s,a)$.  We can estimate the expected discounted return of the policy $\pi_m$ WDR from above; we again perform gradient descent on the gating parameters (with a minibatch of 256 samples).  Due to the nonconvexity of the WDR-based objective, we take the best of $1000$ random restarts.

\section*{Results}

The estimates of the discounted expected return for each policy are presented in Table \ref{tbl:wdr}.  We provide two columns for the mixture of experts policy because it is challenging to derive accurate value estimates $\hat{V}$ and $\hat{Q}$ for the WDR estimator in this case.  Thus, we consider two sensible options: using the value estimates $\hat{V}$ and $\hat{Q}$ from the clinician policy, and using the estimates from the DDQN policy. 

\begin{figure}[ht]
\centering
\includegraphics[width=0.6\textwidth]{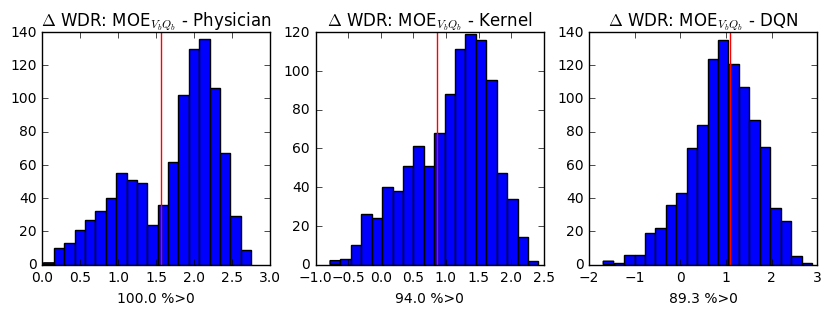}
\caption{The distribution of the WDR estimator difference between MoE and three other policies based on 1000 bootstrapped test datasets. The red vertical lines indicate the difference calculated for the original test dataset.\label{fig:gradients}}
\end{figure}

Regardless of the choice of evaluation covariate, both kernel and DQN policy improve over the physician policy, and the MoE policy projects a further improvement.  Using a recurrent state representation that compresses the entire history results in a further improvement for all RL policies (we used a sparse autoencoder\cite{ng2011cs294a} for the non-recurrent encoding).  The Figure \ref{fig:gradients} includes boostrap intervals for these values\cite{DBLP:journals/corr/abs-1805-12298}.

On average, the WDR estimator predicts the MoE policy to outperform all other proposed policies. We note, however, that for some bootstraped datasets the difference is negative. This is to be expected, as Gottesman et al. [24] demonstrate that high variability of IS based estimators is common in healthcare-data. Gottesman et al. [24] further demonstrate that IS based estimators can have a high selection bias when evaluating policies which are significantly different from the behavior policy.

\begin{table}[H]
\centering
\caption{Estimate of the discounted expected return for policies over test set, $\gamma=0.99$. $V_{d}$ indicates approximating the MoE $V$ by DQN $V$ function, $V_{b}$ indicates approximating the MoE $V$ by behavioral policy, namely, physician $V$ function} 
\label{tbl:wdr}
\begin{tabular}{|c|c|c|c|c|c|}
\hline
       & Physician & Kernel & DQN  &  $MoE_{V_{d}, Q_{d}}$ & $MoE_{V_{b}, Q_{b}}$ \\ \hline
non-recurrent encoded & 3.76 & 3.73  & 4.06 & 3.93 & 4.31 \\ \hline
recurrent encoded & 3.76   & 4.46  & 4.23 & 5.03 & 5.72 \\ \hline
\end{tabular}
\end{table}

\begin{figure*}
  \centering
  \begin{subfigure}[b]{0.5\textwidth}
  \centering
  \includegraphics[width=\linewidth]{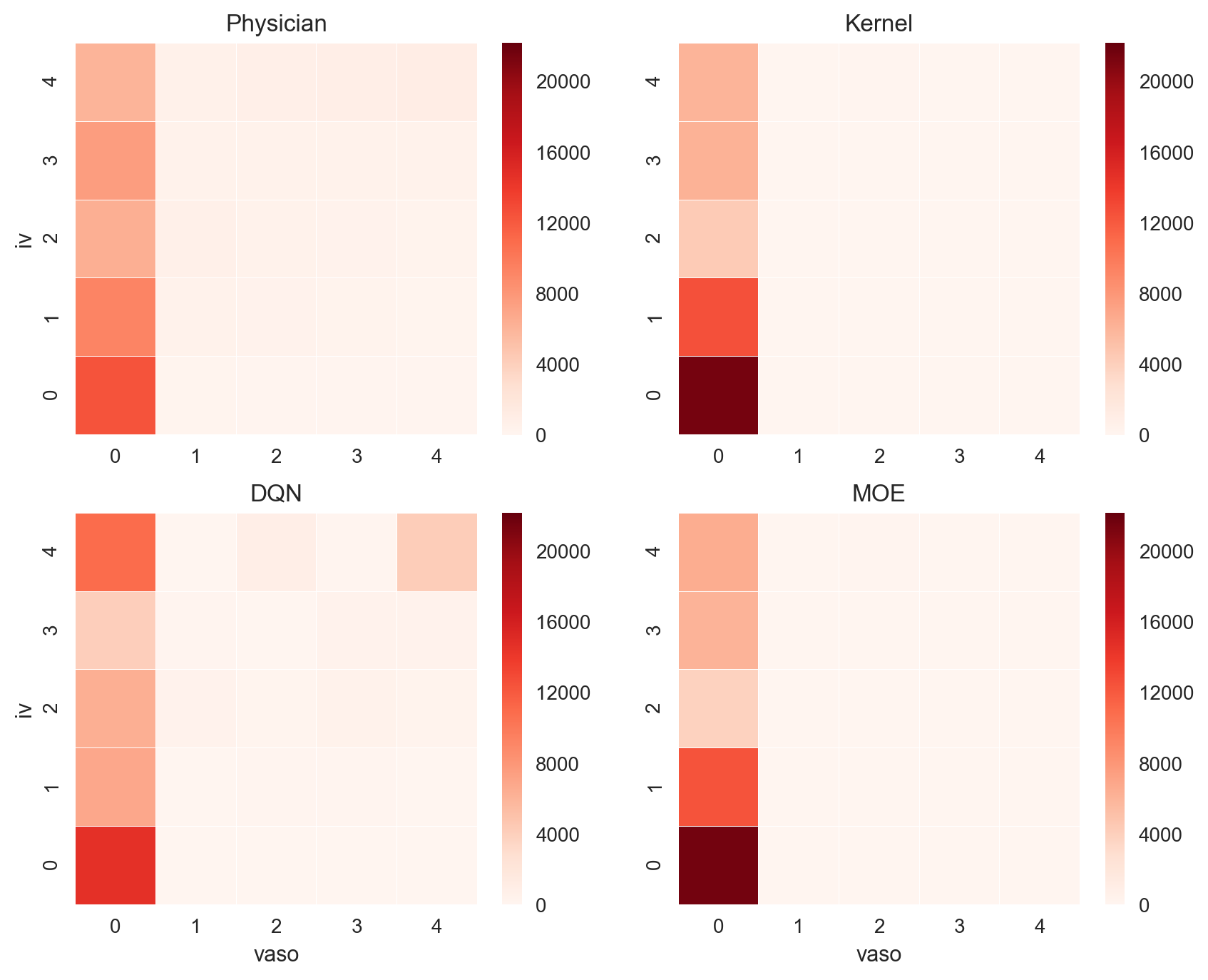}\hfill
  \caption{\label{fig:expert_actions}Action Distributions for physician and experts over test set}
  \label{fig:twelve:b}
  \end{subfigure}%
  \begin{subfigure}[b]{0.5\textwidth}
  \centering
  \includegraphics[width=\linewidth]{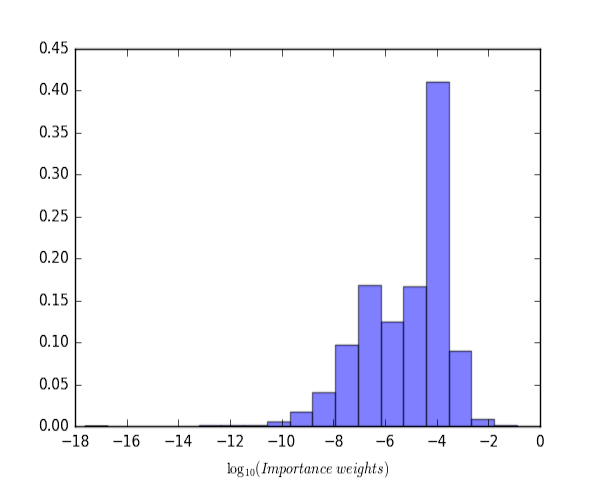}\hfill
  \caption{\label{fig:WDR_weights}WDR importance weights distribution}
  \label{fig:twelve:a}
  \end{subfigure}%
  \caption{The left heatmap shows action distribution of each expert. Action assignment starts at bottom left corner (action 0) in the grid and increases from left to right, with action 24 (max level of fluid and vasopressor) corresponding to the top right corner. The MoE is much akin to the conservative kernel expert, suggesting most of the patients can find similar neighbors. The right chart plots the full distribution of importance weights. A significant number of weights lie in the range of $[10^{-4},10^{-3}]$ and only very few observations have weights significantly larger than that range.}
  \label{fig:twelve}
\end{figure*}

\textit{Analysis of discovered policies.}
Figure \ref{fig:expert_actions} shows the action distributions for the KRL, DRL, clinician, and MoE policies over the test set.  The no treatment $a=0$ action dominates policies.  Actions which are favored by physicians, such as IV but no vasopressor are also (as expected) favored by the kernel policy. That said, the kernel policy tends to be more conservative than the clinicians as it suggests nonaction at approximately twice the clinicians' frequency---perhaps reflecting a bias toward the fact that those patients who were not treated were somehow healthier and thus survived. Perhaps in a similar vein, the kernel expert prescribes more fluid alone than the clinicians: while both suggest high fluids and no vasopressor almost in the same frequency, the kernel expert very rarely suggests actions with vasopressor. 

The DRL policy, like clinician and the kernel policies, favors giving actions a range of fluid values. But, over the test set, DQN expert prescribes more extreme values than clinicians.  Clinicians frequently prescribe high fluids and low vasopressor; however, the DQN policy also tends to give more high vasopressor dosage actions in addition to fluids. 

Over the test set, the MoE policy is closer to the kernel policy. But influenced by the DQN policy, MoE prescribes more high dosage actions.Table~\ref{tbl:overlap} shows how actions suggested by the experts overlap; the rate is high for the kernel and MoE policies again reflecting the fact that most of the time, patients can find similar neighbors.  In 4.4\% of circumstances, the gating results in a MoE policy follows neither that of kernel nor that of DQN policy.   

\begin{table}[H]
\centering
\caption{Percent similarity of different policies over test set patient states}
\label{tbl:overlap}
\begin{tabular}{|c|c|c|c|c|}
\hline
           & kernel & DQN   & MoE   \\ \hline
physician  & 0.305  & 0.151 & 0.296 \\ \hline
kernel     & -      & 0.182 & 0.871 \\ \hline
DQN        & -      & -     & 0.258 \\ \hline
\end{tabular}
\end{table}

\textit{Evaluation Quality Assessment.}
The WDR estimator of policy quality relies on having a large enough collection of patient histories in the evaluation set having non-zero weight $w_t^i$.  For the MoE policy, $90\%$ of the importance sampling weights are non-zero and $86\%$ final weights in the sequences are non-zero. These high numbers of non-zero importance weights indicates that nearly all of our data was used in the evaluation of the policy.  We plot the full distribution of weights in Figure \ref{fig:WDR_weights}. A significant number of weights lie in the range of $[10^{-4},10^{-3}]$ and only very few observations have weights significantly larger than that range (the samples with significantly smaller weights are unlikely to have a significant influence on the estimate). However, the few observations with weights on the order of $10^{-1}$ could potentially have large influence (see Figure \ref{fig:gradients} for variances computed via boostrap).

\textit{Running Time.} 
Besides the quality of policy, the ability to make recommendations quickly is also important in the ICU (note that it is less important for the initial training time to be fast).  We measured the computational time for all components in our framework on a dual-core Intel Core i5 processor. The 2-layer NN for the reward function $f(o)$ took 56.8s to train by 128-sized mini-batch for 50 epochs on dataset with $39856 \times 45$ dimension.  The recurrent autoencoder for the state space took 491s for 50 epochs with $128$ patients per mini-batch.  

The kernel policy required 909s to identify policies for the $150720$ samples in the training set. Training DQN by sampling $30$ transitions for $200,000$ times required $5.15\times 10^4$s. Finally, MoE gating function took 62s to train by a 256-sized mini-batch, for 1 epoch using $1e-4$ learning rate, on dataset with $150720 \times 9$ dimension; however, since MoE cannot guarantee global maxima, we conducted $1000$ random restarts over the initial parameters, and trained each for $50$ epochs. 

Most importantly, with regard to prediction at test time, for a patient with $10$ timesteps in the ICU, it took only 0.162s to encode all the observations, compute the kernel and DRL policies, and compute the gating function for the final MoE policy.

\section*{Discussion}
\label{section:discussion}
Overall, the DQN policy recommended a treatment strategy with more aggressive use of both vasopressors and fluids.   In comparison to the physician policy, DQN recommended  70\% more actions involving medium-to-high fluid volume and vasopressor dosage (actions 18,19,23, and 24). Most notably, frequency for the DQN action corresponding to maximum levels of both fluid and vasopressor (action 24) increased by 3.8 fold from the physician policy.  These results suggest that despite the recent advances in deep reinforcement learning, further investigations are required, and careful clinical judgment should be exercised to guard against potentially high-risk actions introduced from pathologies in non-linear function approximation.   

The proposed kernel policy displayed a different kind of bias.  It recommended far fewer actions involving vasopressors in comparison to both the physician policy and DQN, perhaps because amongst a patient's neighbors, the survivors were relatively healthier and thus treated less aggressively.  By focusing on survivors the kernel policy also focuses on patients who did not just receive a good treatment or were potentially healthier now, but also patients that received good treatments and remained healthy in the future.  If healthier patients are easier to treat in general, then we might expect a bias toward less aggressive treatment from the kernel policy as well.

More broadly, while it appears that our MoE policy significantly outperforms the clinician policy (as well as each individual expert), and we have ensured that the actions it suggests are at least sensible (that is, often taken by clinicians), there still exist a number of limitations.  When encoding the patient clinical course in a recurrent representation, in spite of our high prediction accuracy, we cannot be certain with only these 50 measures that there are no hidden confounding factors; aside from pre-ICU fluid balance, we have no information from prior to their ICU stay.  The choice of representation also influences the quality of our off-policy evaluation, as the WDR estimator assumes that the system in Markov in the state.  More generally, the WDR estimator requires either the behavior policy estimate $\pi_b$ to be accurate or the control variate estimates $\hat{V}$ and $\hat{Q}$ to be accurate to be unbiased; something that we could not guarantee.  That said, we do demonstrate that our results were at least insenstive to different choices of $\hat{V}$ and $\hat{Q}$.

Our work also focused on a very specific reward structure.  To apply off-policy evaluation (WDR) in a statistically credible way, we considered the accumulation of low mortality risk as the objective, rather than mortality itself (as using the latter, the policies could not be evaluated reliably).  To maximize the interpretability of the reward to clinicians, this risk was calculated only from the current observations and not the patient's entire history.  Creating reward functions that can both be checked by human experts and accurately convey clinical goals is a direction for future work. 

Finally, with respect to sepsis management, there also exist many other interventions, such as antibiotics use and mechanical ventilation, that also affect patient outcomes.  Future work remains to investigate policies that incorporated a broader scope of patient history as well as a larger variety of interventions.

\section*{Conclusion}

We presented a MoE framework to learn improved fluid and vasopressor administration strategies for sepsis patients in ICUs using observational data.   We demonstrated that the proposed mixture model approach can automatically adapt to patient states at each time step, and dynamically switch between a conservative kernel policy and a more aggressive deep-RL policy to achieve, under our reward measure, better expected outcomes than clinician, kernel policy, and deep-RL policy. While much further investigation is required to truly validate the efficacy of derived policies, the proposed MoE framework represents a novel approach to take advantage of the strengths of different treatment policies.

\section*{Acknowledgments}

We would like to thank the other students in Harvard CS282R - Reinforcement Learning for Healthcare,  Fall 2017 for their insights, encouragement and feedback. Omer Gottesman was supported by the Harvard Data Science Initiative. L. Lehman was supported by NIH grant 2RO1GM104987.

\makeatletter
\renewcommand{\@biblabel}[1]{\hfill #1.}
\makeatother

\bibliographystyle{unsrt}
\bibliography{reference}




\end{document}